%% file: main.tex
\documentclass[letterpaper]{article} % DO NOT CHANGE THIS
\usepackage{aaai25}  % DO NOT CHANGE THIS

\usepackage{times}  % DO NOT CHANGE THIS
\usepackage{helvet}  % DO NOT CHANGE THIS
\usepackage{courier}  % DO NOT CHANGE THIS
\usepackage[hyphens]{url}  % DO NOT CHANGE THIS
\usepackage{graphicx} % DO NOT CHANGE THIS
\usepackage{natbib}  % DO NOT CHANGE THIS AND DO NOT ADD ANY OPTIONS TO IT
\usepackage{caption} % DO NOT CHANGE THIS AND DO NOT ADD ANY OPTIONS TO IT
\usepackage{algorithm}
\usepackage{algorithmic}

\usepackage[table]{xcolor}

\usepackage{latexsym}
\usepackage{amssymb}
\usepackage{amsmath}
\usepackage{amsthm}
\usepackage{booktabs}
\usepackage{enumitem}
\usepackage{graphicx}
\usepackage{color}
\usepackage{overpic}
\usepackage{multirow}
\usepackage{xcolor}         % colors
\usepackage{colortbl}
\usepackage{bbding}
\usepackage{pifont}
\usepackage{fontawesome}
\usepackage{graphics}
\usepackage{lipsum}

\definecolor{lightgrey}{RGB}{244,244,244}
\definecolor{grey}{RGB}{128,128,128}
\definecolor{midgrey}{RGB}{225,225,225}
\definecolor{forestgreen}{RGB}{47, 159, 87}

\usepackage[capitalize]{cleveref}
\crefname{section}{Sec.}{Secs.}
\Crefname{section}{Section}{Sections}
\crefname{table}{Tab.}{Tabs.}
\Crefname{table}{Table}{Tables}

\usepackage{soul}
\graphicspath{{figures/}}

\definecolor{Gray}{gray}{0.95}
\definecolor{Cyan}{rgb}{0.88,1,1}

\usepackage{newfloat}
\usepackage{listings}
\DeclareCaptionStyle{ruled}{labelfont=normalfont,labelsep=colon,strut=off} % DO NOT CHANGE THIS
\floatstyle{ruled}
\newfloat{listing}{tb}{lst}{}
\floatname{listing}{Listing}
\nocopyright

\title{LLM4GEN: Leveraging Semantic Representation of LLMs \\ for Text-to-Image Generation}
\author {
    Mushui Liu$^{1,}$\thanks{Equal Contribution.}, Yuhang Ma$^{2,*}$, Zhen Yang$^{1}$, Jun Dan$^{1}$, \\ Yunlong Yu$^{1,}$\thanks{Co-Corresponding Author}, Zeng Zhao$^{2,\dagger}$, Zhipeng Hu$^{2}$, Bai Liu$^{2}$, Changjie Fan$^{2}$
}
\affiliations{
    $^{1}$Zhejiang University, $^2$ Fuxi AI Lab, Netease Inc.
}

\usepackage{bibentry}
\begin{document}

\maketitle

\input{sections/abstract}

\input{sections/introduction}

\input{sections/related_work}

\input{sections/method}
\input{sections/experiment}

\input{sections/conclusion}

\bibliography{aaai25}

\end{document}

%% file: sections/abstract.tex
\begin{abstract}
\noindent
Diffusion models have exhibited substantial success in text-to-image generation. However, they often encounter challenges when dealing with complex and dense prompts involving multiple objects, attribute binding, and long descriptions. In this paper, we propose a novel framework called \textbf{LLM4GEN}, which enhances the semantic understanding of text-to-image diffusion models by leveraging the representation of Large Language Models (LLMs). It can be seamlessly incorporated into various diffusion models as a plug-and-play component. A specially designed Cross-Adapter Module (CAM) integrates the original text features of text-to-image models with LLM features, thereby enhancing text-to-image generation. Additionally, to facilitate and correct entity-attribute relationships in text prompts, we develop an entity-guided regularization loss to further improve generation performance. We also introduce DensePrompts, which contains $7,000$ dense prompts to provide a comprehensive evaluation for the text-to-image generation task. Experiments indicate that LLM4GEN significantly improves the semantic alignment of SD1.5 and SDXL, demonstrating increases of 9.69\% and 12.90\% in color on T2I-CompBench, respectively. Moreover, it surpasses existing models in terms of sample quality, image-text alignment, and human evaluation.
\end{abstract}

%% file: sections/introduction.tex
\section{Introduction}

Recently, diffusion models \cite{song2020score,rombach2022high,chen2023controlstyle} have made significant progress in text-to-image (T2I) generation models, such as Imagen \citep{saharia2022photorealistic}, DALL-E \cite{betker2023improving}, and Stable Diffusion \cite{rombach2022high, podell2023sdxl}. However, they often encounter challenges in generating images given complex and dense prompt descriptions, such as attribute binding and multiple objects \cite{huang2023t2i}.

\begin{figure}[!t]
    \centering
    \includegraphics[width=\linewidth]{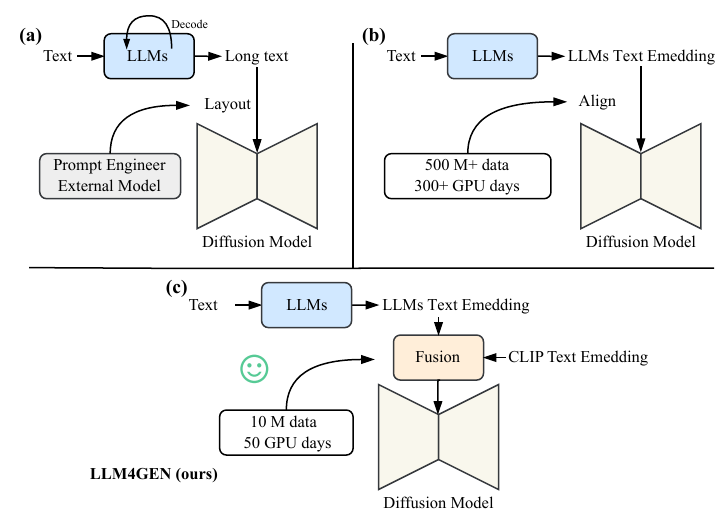}
    \caption{Architecture comparison between (a) LLM-guidance models (b) LLM-alignment models and (c) our proposed LLM4GEN.}
    \vspace{-2mm}
    \label{fig:intro-framework}
\end{figure}

With the emergence of powerful linguistic representations from Large Language Models (LLMs), there has been an increasing trend in leveraging LLMs to aid in T2I generation.
Current methods mainly consist of two categories: LLM-guidance models \citep{RPG,feng2022training} and LLM-alignment models \citep{wu2023paragraph, ella, sd3,bridge}.
LLM-guidance models harness the reasoning capability of LLMs and the Layout model to generate controllable images, as illustrated in \cref{fig:intro-framework}~(a).
However, these methods require separating LLMs from external models, resulting in redundancy in both inference time and the overall framework. While LLM-alignment models utilize LLMs to exploit the representational capacity, they demand substantial training data to align LLM representations with the diffusion model, as shown in \cref{fig:intro-framework}~(b).

\begin{figure}[!t]
    \centering
   \includegraphics[width=\linewidth]{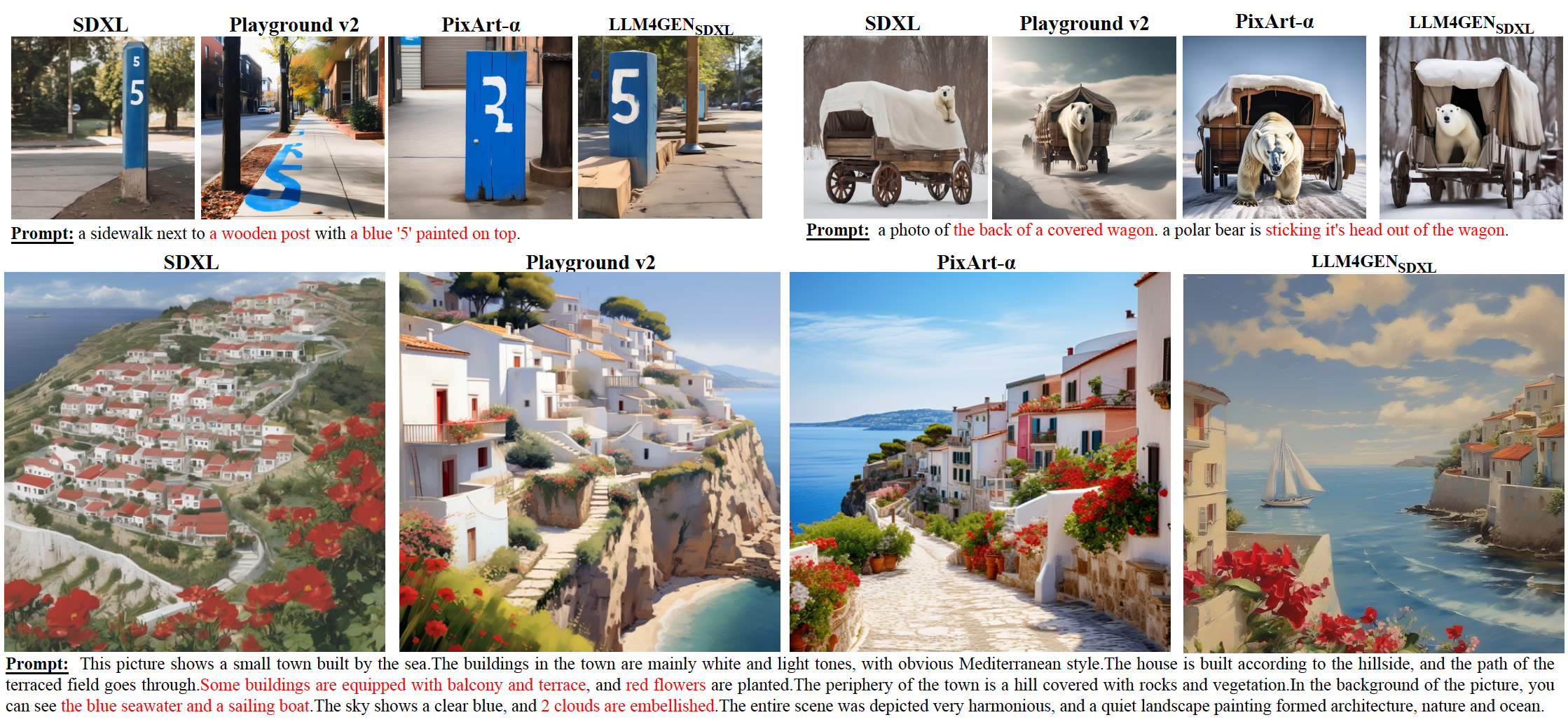}
  \caption{Image generation using concise and dense prompts, with colored text highlighting key entities or attributes(Zoom in for details).}
    \vspace{-2mm}
  \label{fig:intro-1}
\end{figure}

To address aforementioned challenges, we propose \textbf{LLM4GEN}, a novel framework that implicitly leverages the powerful semantic representations of \textbf{LLMs} to enhance the original text encoder \textbf{for} T2I \textbf{GEN}eration, as illustrated in \cref{fig:intro-framework}~(c).
Specifically, we design an efficient Cross-Adapter Module (CAM) to implicitly integrate the semantic representation of LLMs with original text encoders that have limited representational capabilities, such as CLIP text encoder \cite{clip}.
We apply cross-attention to the representations of both encoder-only LLMs (e.g., Llama \cite{zhang2023llama}) and decoder-only LLMs (e.g., T5 \cite{raffel2020exploring}) alongside CLIP text embeddings, and then concatenate the fused embedding with the original CLIP text embedding.
This CAM module significantly enhances the performance of T2I diffusion models while preserving the original text encoder representations, thereby reducing the need for extensive training data.
Additionally, we introduce an entity-guidance regularization loss that penalizes mismatches between the activation maps of entities and their corresponding attributes in the text, improving the model's ability to accurately comprehend and represent the main subjects in the generated images.
As evidenced in \cref{fig:intro-1}, our proposed method exhibits strong performance in T2I generation.

To comprehensively assess the image generation capabilities of T2I models, we develop a comprehensive benchmark named \textbf{DensePrompts}, an extension of T2I-CompBench \cite{huang2023t2i}, which incorporates over 7,000 compositional prompts.
The construction of this benchmark involves leveraging LLMs for complex text descriptions, followed by manual refinement. Results from performance metrics and human evaluations consistently demonstrate that LLM4GEN's representational capability surpasses other existing methods.

Overall, our contributions are as follows: 
\begin{itemize}
    \item We propose a novel framework that leverages the powerful representational capabilities of LLMs to assist in text-to-image (T2I) generation. Specifically, we design a Cross-Adapter to integrate LLM representations and introduce an entity-guidance regularization loss to enhance semantic understanding.
    \item To assess performance with long-text prompts, we introduce DensePrompts, a benchmark designed to evaluate both aesthetic quality and image-text alignment.
    \item Our designed LLM4GEN can be seamlessly integrated into existing diffusion models like SD1.5 \cite{rombach2022high} and SDXL \cite{podell2023sdxl}. Experiments show that LLM4GEN exhibits superior performance in sample quality, image-text alignment, and human evaluation compared with existing state-of-the-art models.
\end{itemize}

%% file: sections/related_work.tex
\section{Related Work} \label{sec:related}
\textbf{Large Language Models}

Large language models (LLMs) \cite{chang2023survey} have shown powerful generalization ability in various NLP tasks. Recent LLMs, e.g., GPTs \cite{brown2020language}, LLaMA \cite{touvron2023llama}, OPT \cite{zhang2022opt}, PaLM \cite{chowdhery2022palm} are all equipped with billions of parameters, enabling the intriguing capability for in-context learning and demonstrating excellent zero-shot performance across various tasks. Certain Multi-modal LLMs (MLLMs) \cite{gpt-4,zhu2023minigpt,bai2023qwen} have integrated visual and audio modalities, enhancing intelligent interactions with the help of LLMs. \cite{pang2023frozen} shows that the frozen LLMs can further integrate visual understanding. Recent works \cite{lmd,emu,RPG} use LLMs to create improved text prompts or bounding box layouts for high-quality text-to-image generation. However, these existing works only consider LLMs as simple condition generators, e.g., text prompts or layout planning. In this paper, we harness the representation capabilities of LLMs to enhance text-to-image generation, emphasizing their significant representational power beyond simple text output.

\noindent
\textbf{Text-to-Image Diffusion Models} 
Text-to-image generation aims to create images with given prompts. Diffusion models \cite{song2019generative,song2020improved,song2020score} have demonstrated remarkable performance in image generation. These models use added Gaussian noise for a forward process and can generate diverse, high-quality images through an inverse process from random Gaussian noise. GLIDE \cite{nichol2021glide} utilizes CLIP \cite{clip} text encoder to enhance the image-text alignment. Latent Diffusion Models (LDMs) \cite{rombach2022high} transfer the diffusion process from pixel to latent space. Recent models such as SD-XL \cite{podell2023sdxl}, DALL-E 3 \cite{betker2023improving}, and Dreambooth \cite{ruiz2023dreambooth} have significantly enhanced image quality and text-image alignment using various perspectives, such as training strategies and scaling training data. Despite these notable advancements, generating high-fidelity images aligned with complex textual prompts remains challenging. In this paper, we propose LLM4GEN, which leverages the robust representation capabilities of LLMs to facilitate image generation from textual descriptions.

\begin{figure*}[!ht]
    \centering
    \includegraphics[width=\linewidth]{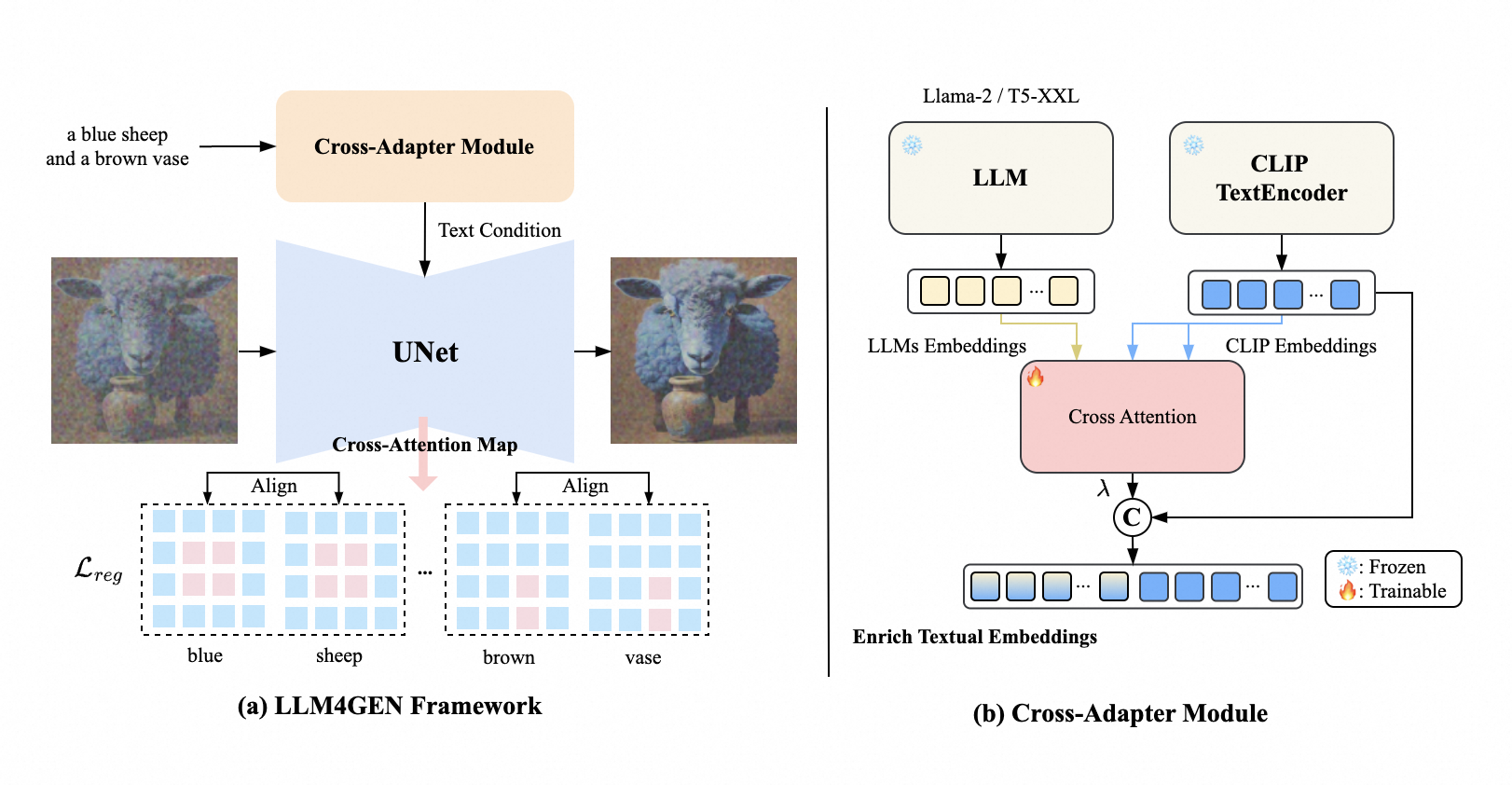}
    \vspace{-10mm}
    \caption{The overview of LLM4GEN. (a) Framework. (b) Cross-Adapter Module.}
    \label{fig:framework}
\end{figure*}

%% file: sections/method.tex
\section{Methodology}

\subsection{LLM4GEN} \label{sec:framework}
\subsubsection{Framework}
The proposed LLM4GEN, which contains a Cross-Adapter Module (CAM) and the UNet, is illustrated in \cref{fig:framework}~(a). In this paper, we explore stable diffusion \cite{rombach2022high, podell2023sdxl} as the base text-to-image diffusion model, and the vanilla text encoder is from CLIP \cite{clip}. LLM4GEN leverages the strong capability of LLMs to assist in text-to-image generation. The CAM extracts the representation of a given prompt via the combination of LLM and CLIP text encoder. The fused text embedding is enhanced by leveraging the pre-trained knowledge of LLMs through the simple yet effective CAM. By feeding the fused text embedding, LLM4GEN iteratively denoises the latent vectors with the UNet and decodes the final vector into an image with the VAE. 

\subsubsection{Cross-Adapter Module}
The CAM connects the LLMs and the CLIP text encoder using a cross-attention layer, followed by concatenation with the representation of the CLIP text encoder. The last hidden state of the LLMs is extracted as LLMs feature $c_{l}$. The feature of CLIP text encoder is denoted as $c_{t}$, and we perform a cross-attention to fuse them:
\begin{equation}
\label{eq:cross-adapter-expanded}
\begin{aligned}
Q=W_q(c_{l}), K=W_k(c_{t}), V=W_v(c_{t}) 
\end{aligned}
\end{equation}
\begin{equation}
\label{eq:cross-adapter-expanded}
\begin{aligned}
c_{l}' = \operatorname{CrossAttention}(Q, K, V)=\operatorname{softmax}\left(Q \cdot K^T\right) \cdot V
\end{aligned}
\end{equation}
where $W_q$, $W_k$, $W_v$ are the trainable linear projection layers. The output embedding dimension is the same as that of CLIP text encoder. Then the final fused text embedding of the CAM is:
\begin{equation}
\label{eq:unet-cross-attention}
\begin{aligned}
x = \operatorname{CA}(x, \operatorname{Concat}(\lambda \cdot c_{l}', c_{t})) = \lambda \cdot \operatorname{CA}(x, c_{l}) + \operatorname{CA}(x, c_{t})
\end{aligned}
\end{equation}
where $\operatorname{Concat}$ denotes concatenation in the sequence dimension, and $\lambda$ is the balance factor, $x$ denotes the latent noise, $\operatorname{CA}$ is the cross-attention module within the UNet module. Overall, our designed Cross-Adapter Module implicitly facilitates the strong representation of LLMs with a residual fusion manner, without utilizing extensive training data and resources to condition the latent vectors on text embeddings. Notably, our LLM4GEN is compatible with both decoder-only and encoder-only LLMs and we evaluate on Llama-2 7B/13B \cite{touvron2023llama} and T5-XL \cite{brown2020language} in further experiments. 

\begin{figure}[!ht]
\centering
\includegraphics[width=0.8\linewidth]{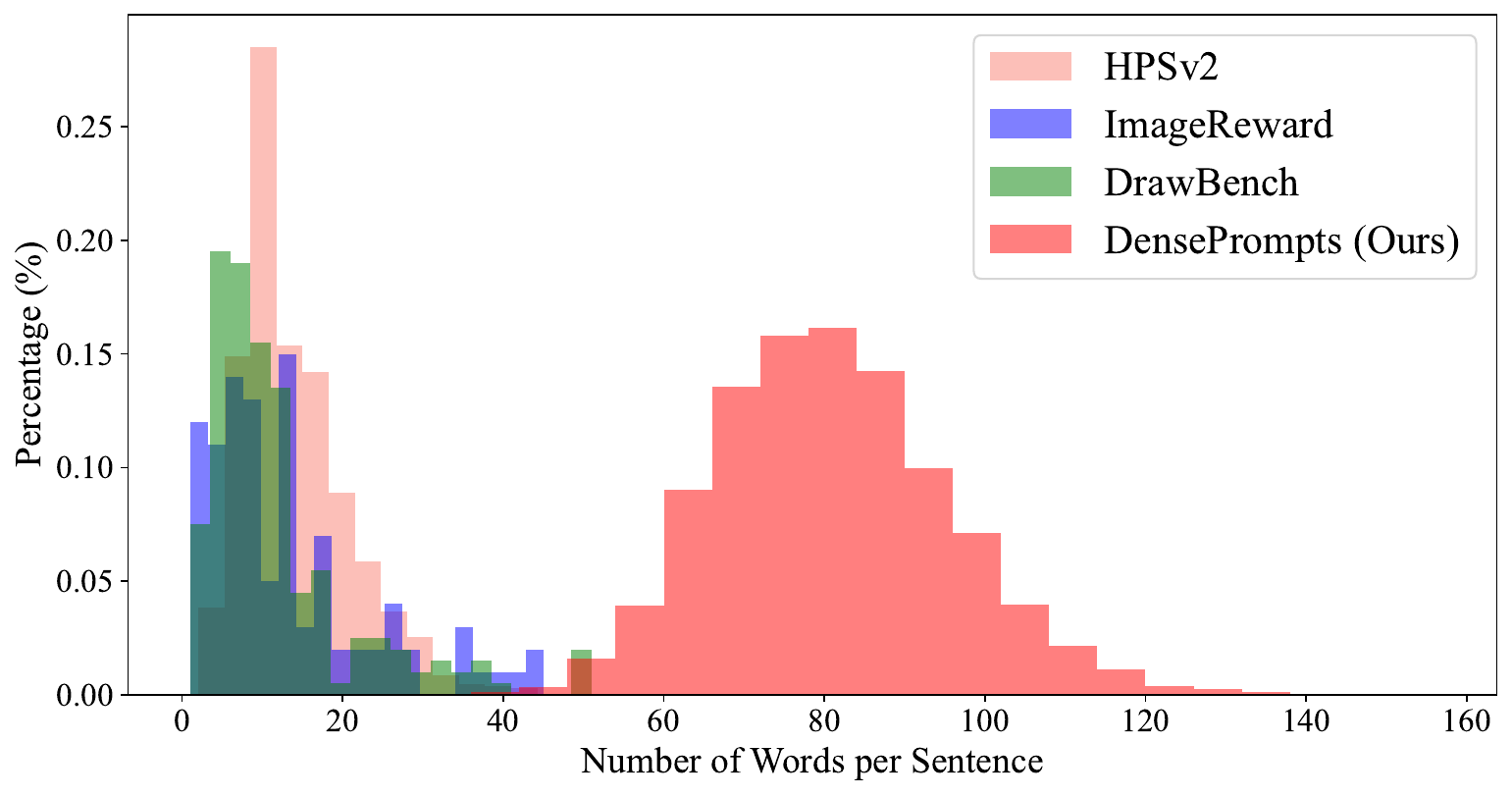}
\caption{Statistic of DensePrompts benchmark compared with other benchmarks.}
\label{fig:statistic}
\end{figure}

\subsubsection{Entity-Guidance Regularization Loss}
Current text-to-image generation models often encounter confusion and omissions when generating multiple entities. 
We utilize a parser to analyze the prompt $\mathcal{P}$, extracting a set of attribute-entity pairs $\mathcal{S} = \sum_{i=1}^N \{a_{i}, e_{i}\}$, where $e_{i}$ and $a_{i}$ represent the entity name and its corresponding attribute, respectively, and $N$ denotes the number of parsed pairs. Subsequently, we can calculate the active similarity map as:
\begin{equation} \label{eq: activation map}
    \mathcal{A}_{i} = \operatorname{softmax}\left(\frac{QK_{i}^T}{\sqrt{d}}\right)
\end{equation}
where the query $Q$ is derived from the latent representation, the key $K_{i}$ is derived from the token embedding of $p$, and $d$ is the latent dimension. $\mathcal{A}_{a}$ and $\mathcal{A}_{o}$ indicate the similarity maps for the attribute word and the entity, respectively. Subsequently, we impose a penalty on these similarity maps on all UNet layers as:
\begin{equation} \label{eq: sim}
    \mathcal{L}_{reg} = \frac{1}{N\cdot L}\sum_{i=1}^N\sum_{l=1}^L||\mathcal{A}_{a}^{i-l} - \mathcal{A}_{o}^{i-l}||^{2}
\end{equation}
where $||\cdot||$ represents the L2 distance, $L$ is the layer numbers.

Overall, based on the framework described above, the training loss of LLM4GEN is formulated as:
\begin{equation}
\label{eq:training}
\mathcal{L} = \mathbb{E}_{\varepsilon(x),\epsilon\sim \mathcal{N}(0,1),t}[\parallel\epsilon-\epsilon_{\theta}(z_{t},t)\parallel]_{2}^{2}] + \alpha \cdot \mathcal{L}_{reg}
\end{equation}
where $z_{t}$ can be obtained from the encoder $\mathcal{E}$, ans latent vectors from $p(z)$ can be decoded to images through the decoder $\mathcal{D}$. In this paper, we address the limited representation of CLIP as a text encoder by leveraging the capabilities of large language models (LLMs) to enhance the text encoder of the LDMs.

\subsection{DensePrompts Benchmark} 
A comprehensive benchmark is crucial for evaluating the image-text alignment of generated images. Current benchmarks, e.g., MSCOCO \cite{microsoftcoco} and T2I-CompBench \cite{huang2023t2i}, primarily consist of concise textual descriptions, are not comprehensive enough to describe a diverse range of objects. Thus, we introduce a new comprehensive and complicated benchmark called \textbf{DensePrompts}, comprising lengthy textual descriptions.

Initially, we collect 100 images from the Internet, comprising 50 real and 50 generated images, each with intricate details. Leveraging the robust image comprehension capabilities of GPT-4V \cite{2023GPT4VisionSC}, we utilize it to provide detailed descriptions for these 100 images, encompassing object attributes and their relationships, thereby generating comprehensive prompts abundant in semantic details. We employ GPT-4 \cite{gpt-4} to produce massive long texts based on generated prompts mentioned above. DensePrompts provides more than 7,000 extensive prompts whose average word length is more than 40. Word statistics of DensePrompts are outlined in \cref{fig:statistic}. To assess the performance, DensePrompts benchmark incorporates CLIP Score \cite{clip} and Aesthetic Score \cite{improvedaestheticpredictor}. Combining our proposed DensePrompts with T2I-CompBench, we establish a comprehensive evaluation in text-to-image generation.

%% file: sections/experiment.tex
\section{Experiments} \label{sec:experiments}
\subsection{Experimental Details}
\subsubsection{Framework and Implementation Details} In this paper, we explore LLM4GEN based on SD1.5 and SDXL, denoted as LLM4GEN$_{SD1.5}$ and LLM4GEN$_{SDXL}$. We utilize T5-XL and CLIP text encoder (CLIP ViT-L/14) as the text tower. The sequence length of the LLMs is set to $128$. We use 10M text-image pairs collected from LAION-2B\cite{laion} and Internet. Training is conducted on 8NVIDIA A100 GPUs with the learning rates of 2e-5 and 1e-5 for LLM4GEN$_{SD1.5}$ and LLM4GEN$_{SDXL}$, respectively. The batch size is set to 256 and 128. The training steps are set to 20k and 40k. Additionally, we further train LLM4GEN$_{SDXL}$ using 2M high-quality data with 1024 resolution. During inference, we utilize DDIM sampler \cite{ddim} for sampling with 50 steps and the classifier free guidance scale to 7.5.

\subsubsection{Evaluation Benchmarks} We comprehensively evaluate proposed LLM4GEN via four primary benchmarks, \textit{e.g.} \textbf{MSCOCO} \cite{microsoftcoco}, \textbf{T2I-CompBench} \cite{huang2023t2i}, our proposed \textbf{DensePrompts} benchmark, and \textbf{User Study}.

\begin{table*}[!ht]
    \centering
    \begin{tabular}{lcccccc}
    \toprule
    \multirow{2}{*}{Model} & \multicolumn{3}{c}{Attribute Binding} & \multicolumn{2}{c}{Object Relationship} & \multirow{2}{*}{Complex$\uparrow$} \\
    & Color $\uparrow$ & Shape $\uparrow$ & Texture $\uparrow$ & Spatial $\uparrow$ & Non-Spatial $\uparrow$ & \\
    \midrule
    Composable Diffusion \cite{liu2022compositional} & 40.63 & 32.99 & 36.45 & 8.00 & 29.80 & 28.98 \\
    Structured Diffusion \cite{feng2022training} & 49.90 & 42.18 & 49.00 & 13.86 & 31.11 & 33.55 \\
    Attn-Exct v2 \cite{chefer2023attend} & 64.00 & 45.17 & 59.63 & 14.55 & 31.09 & 34.01 \\
    GORS \cite{huang2023t2i} & 66.03 & 47.85 & 62.87 & 18.15 & 31.93 & 33.28 \\ 
    DALL-E 2 \cite{ramesh2022hierarchical} & 57.50 & 54.64 & 63.74 & 12.83 & 30.43 & 36.96 \\
    PixArt-$\alpha$ \cite{chen2023pixart} & 68.86 & 55.82 & {70.44} & 20.82 & 31.79 & 41.17 \\
    SUR-Adapter\footnotesize{*} \cite{suradapter} & 38.93 & 36.73 & 39.21 & 13.21 & 30.81 & 31.43 \\ 
    ELLA$_{SDXL}$ \cite{ella} & 72.60 & 56.34 & 66.86 & 22.14 & 30.69 & - 
    % lavidge
    \\ 
    % \textcolor{gray}{SD-3.0\footnotesize{*}} & \textcolor{gray}{79.98} & \textcolor{gray}{67.23} & \textcolor{gray}{74.51} & \textcolor{gray}{26.35} & \textcolor{gray}{34.32} & \textcolor{gray}{44.13} \\
    \midrule
    SD1.5 \cite{rombach2022high} & 37.65 & 35.76 & 41.56 & 12.46 & 30.79 & 30.80 \\
    \textbf{LLM4GEN$_{SD1.5}$ (Ours)} & 47.34 & 45.35 & 55.39 & 14.79 & 31.00 & 33.97 \\
    $\Delta$ (Margin) & +9.69 & +9.59 & +13.83 & +2.33 & +0.21 & +3.17 \\
    \midrule
    SDXL \cite{podell2023sdxl} & 63.69 & 54.08 & 56.37 & 20.32 & 31.10 & 40.91 \\
    \textbf{LLM4GEN$_{SDXL}$ (Ours) }& \textbf{76.59} & \textbf{59.24} & \textbf{70.86} & \textbf{24.12} & \textbf{32.10} & \textbf{42.19} \\
    $\Delta$ (Margin) & +12.90 & +5.16 & +14.49 & +3.80 & +1.00 & +1.28 \\
    \bottomrule
    \end{tabular}
    \caption{Evaluation results (\%) on T2I-CompBench \cite{huang2023t2i}. The higher is better, and the best results are highlighted in bold. * denotes that we calculate the metrics using the official weights and code provided.}
    \label{tab:t2i}
\end{table*}

\begin{table}[!t]
  \centering
    \resizebox{\linewidth}{!}{
    \begin{tabular}{lccc}
    \toprule
    \textbf{Method}
      & \textbf{FID}$\downarrow$ & \textbf{IS}$\uparrow$ & \textbf{CLIP Score}(\%)$\uparrow$ \\
    \midrule
    SD1.5 \cite{rombach2022high} & 26.89 & 32.24 & 28.66 \\
    SD1.5 (ft) & 25.48 & 33.53 & 29.10 \\ 
    LLM4GEN$_{SD1.5}$ & \textbf{25.20} & \textbf{34.24} & \textbf{29.45} \\
    \midrule
    SDXL \cite{podell2023sdxl} & 24.75 & 34.91 & 30.10 \\
    LLM4GEN$_{SDXL}$ & \textbf{24.21} & \textbf{35.10} & \textbf{30.91} \\
    \bottomrule
    \end{tabular}
    }
  \caption{Quantitative comparison on text-to-image generation models on the subset of MSCOCO \cite{microsoftcoco} dataset.}
  \vspace{-5mm}
    \label{tab:mscoco}
\end{table}

\begin{figure}[!ht]
    \centering
    \includegraphics[width=0.9\linewidth]{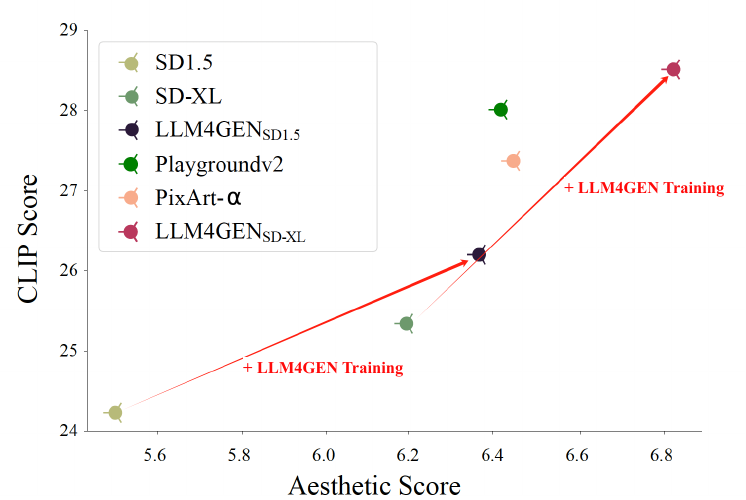}
    \caption{Aesthetic Score and CLIP Score (\%) on DensePrompts benchmark.}
    \vspace{-5mm}
    \label{fig:dense-prompts}
\end{figure}

\begin{figure}[!t]
    \centering
    \includegraphics[width=\linewidth]{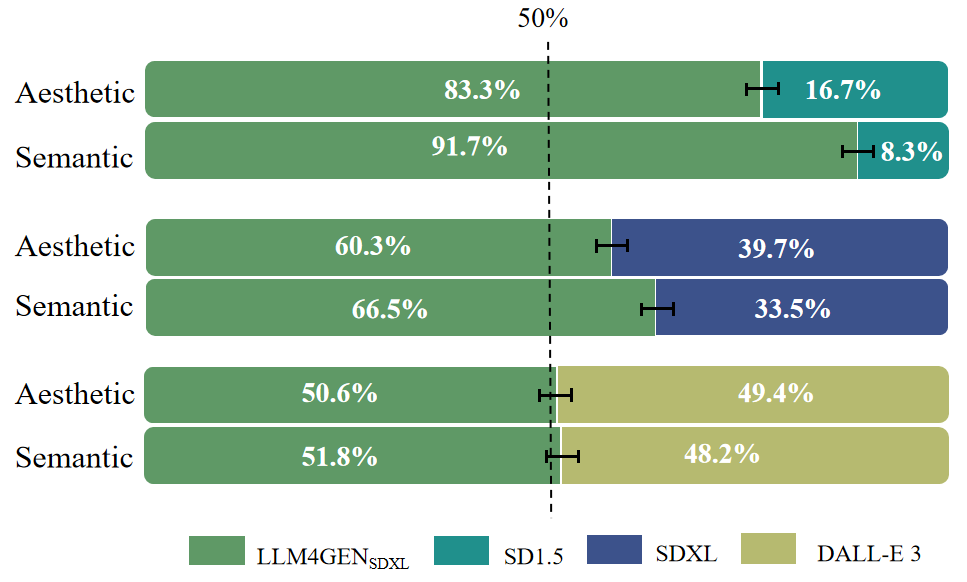}
    \caption{Results on user study regarding the sample quality and image-text alignment of different models.}
    \label{fig:user-study}
\end{figure}

\begin{figure*}[!ht]
    \centering
    \includegraphics[width=\linewidth]{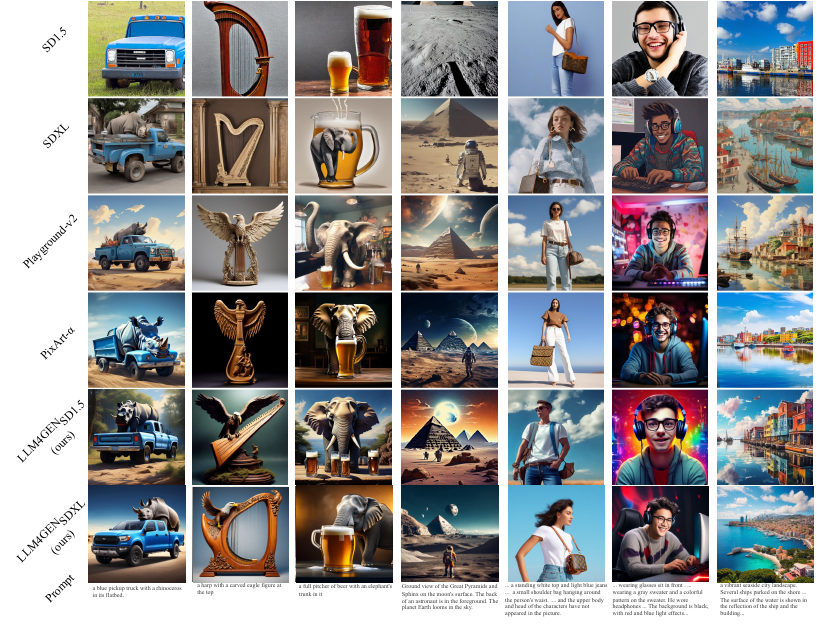}
    \caption{A comparative analysis of LLM4GEN and other state-of-the-art diffusion models using PartiPrompts \cite{yu2022scaling} and our proposed DensePrompts as prompts. The last row represents the prompts used.}
    \label{fig:vis-qual}
\end{figure*}

\subsection{Performance Comparisons and Analysis}
\textbf{Fidelity assessment on MSCOCO benchmark}
Experimental results on MSCOCO benchmark are shown in \cref{tab:mscoco}. LLM4GEN notably enhances the sample quality and image-text alignment, resulting in improvements of 1.79 and 0.54 on FID compared to SD1.5 and SDXL, respectively. Furthermore, we assess the performance of SD1.5 after extensive fine-tuning with the same training dataset. This modified version, SD1.5 (ft), surpasses the original SD1.5, yet LLM4GEN$_{SD1.5}$ still exhibits superior performance over SD1.5 (ft). This underscores the potent representation of our proposed LLM4GEN and its contribution to text-to-image generation.

\noindent
\textbf{Evaluation on T2I-CompBench}
For T2I-CompBench comparison, we select the recent text-to-image generative models for comparison, e.g., Composable Diffusion, Structured Diffusion, Attn-Exct v2, GORS, DALLE 2, PixArt-$\alpha$, ELLA$_{SDXL}$, SD1.5, and SDXL. Experimental results shown in \cref{tab:t2i} demonstrate the distinctive performance of LLM4GEN$_{SDXL}$ in T2I-CompBench evaluation, underlining its advancements in attribute binding, object relationship, and mastery in rendering complex compositions. LLM4GEN shows considerable improvement in color, shape, and texture, showcasing enhancements up to +12.90\% in color, +5.16\% in shape, and +14.49\% in texture with SDXL, respectively. LLM4GEN$_{SDXL}$ also marks considerable progress in both spatial and non-spatial evaluations, with 3.80\% and 1.00\% lift, respectively. Furthermore, when compared with PixArt-$\alpha$, which employs T5-XL as its text encoder, LLM4GEN$_{SDXL}$ surpasses it in several aspects, such as a notable 7.73\% lead in color metric. Moreover, LLM4GEN$_{SDXL}$ outperforms ELLA$_{SDXL}$. These results verify the potent synergy of LLMs representations in augmenting the sample quality and image-text alignment of diffusion models.

\noindent
\textbf{Evaluation on DensePrompts}
We compare our LLM4GEN with PixArt-$\alpha$, Playground v2, SD1.5, and SDXL on our DensePrompts benchmark. As shown in \cref{fig:dense-prompts},  LLM4GEN$_{SDXL}$ achieves the highest Aesthetic Score and CLIP Score among these models. PixArt-$\alpha$ outperforms SDXL due to its T5-XL text encoder for dense prompts. LLM4GEN excels in understanding and interpreting dense prompts, resulting in high-quality images with strong image-text alignment.
This performance is attributed to the powerful representation of LLMs and the effective adaptation of the original CLIP text encoder via our CrossAdapter Module. We attribute this performance to th  powerful representation of LLMs and the effective adaptation of the original CLIP text encoder via our CrossAdapter Module.

\noindent
\textbf{Quantitive Results.} To thoroughly evaluate our proposed LLM4GEN framework, we present the qualitative results on the short prompts provided by PartiPrompts \cite{yu2022scaling} in the first 4 columns and on the dense prompts provided by DensePrompts in the last 3 columns in \cref{fig:vis-qual}. The results indicate that our proposed LLM4GEN$_{SD1.5}$ and LLM4GEN$_{SDXL}$ exhibit strong text-image alignment and superior dense prompt generation compared to the recent PixArt-$\alpha$, especially in handling the multiple objects and attribute binding.

\noindent
\textbf{User Study} We conduct the user study on various combinations of existing methods and LLM4GEN$_{SDXL}$. For each pairing, we assess two criteria: sample quality and image-text alignment. Users are tasked with evaluating the aesthetic appeal and semantic understanding of images with identical text to determine the superior one based on these assessment criteria. Subsequently, we compute the percentage scores for each model, as shown in \cref{fig:user-study}. The results showcase our LLM4GEN$_{SDXL}$ exhibits comparative advantages over both SD1.5 and SDXL. Specifically, LLM4GEN$_{SDXL}$ achieves 60.3\% and 66.5\% higher voting preferences compared to SDXL in terms of Aesthetic and Semantic, respectively. Notably, LLM4GEN$_{SDXL}$ also delivers competitive results when compared to DALL-E 3. 

\subsection{Ablation Studies}
\subsubsection{Impact of Cross-Adapter Module} 
Due to limited computing sources, we evaluate the impact of various architectural enhancements on SD1.5, as outlined in \cref{tab:ab-design}. Our configurations explore different methods for integrating LLMs embeddings: (1) {the baseline SD1.5 model}, (2) {SD1.5 finetune-CLIP}, the result of fine-tuning the original text-encoder of SD1.5, (3) {MLP or CrossAttention}, which utilizes a simple linear layer or cross-attention layer to transform LLM embeddings, (4) {MLP + Concat}, representing a process where LLMs embeddings are projected to the same dimension as the original text embeddings before concatenation, (5) {CrossAttention + Concat}, (6) {CLIP as Q and LLM as KV}, refering to converting the position of Q and KV in CAM. Results show that configuration (2) indeed brings an improvement of the original SD1.5, yet, due to the limited semantic representation of CLIP, the results still remain subpar. Interestingly, simply concatenating the original text embeddings (configuration 3 \& 4) provides a significant boost over base SD1.5. This suggests that direct representation alignment between LLMs and the latent vector is challenging, and enhancing the original text embeddings with LLM embeddings is sufficient to improve image-text alignment. In our LLM4GEN, the LLM representation is employed as Q, while the original text encoder serves as K and V. We also examine the impact of rearranging the position of Q and KV in the CAM module. The results, as demonstrated in configuration 6, indicate that our LLM4GEN (configuration 7) exceeds it, showcasing a 3.65\% enhancement in color. This emphasizes the substantial benefits of incorporating our Cross-Adapter Module to enrich the representation of the original text encoder and the image-text alignment of generated images.

\begin{table}[!t]
    \centering
    \resizebox{\linewidth}{!}{
    \begin{tabular}{l|ccc}
    \toprule
        \multirow{2}{*}{LLMs} & \multicolumn{3}{c}{Attribute Binding} \\
         & Color $\uparrow$ & Shape $\uparrow$ & Texture $\uparrow$ \\ 
    \midrule
        SD1.5 \cite{rombach2022high} & 37.65 & 35.76 & 41.56\\
        Llama-2/7B \cite{touvron2023llama} & 43.21 & 40.12 & 48.91 \\ 
        Llama-2/13B \cite{touvron2023llama} & 44.98 & 41.03 & 49.21 \\
        T5-XL \cite{raffel2020exploring} & 47.34 & 45.35 & 55.39 \\ 
    \bottomrule
    \end{tabular}
    }
    \caption{Impact (\%) of Different LLMs based on SD1.5.}
    \vspace{-4mm}
    \label{tab:ab-llm}
\end{table}

\begin{table}[!t]
    \centering
    \resizebox{\linewidth}{!}{
    \begin{tabular}{l|ccc}
    \toprule
    \multirow{2}{*}{Module} & \multicolumn{3}{c}{Attribute Binding} \\
    & Color $\uparrow$ & Shape $\uparrow$ & Texture $\uparrow$ \\ 
    \midrule
    (1)~~SD1.5 & 37.65 & 35.76 & 41.56 \\
    (2)~~SD1.5 finetune-CLIP & 38.76 & 36.85 & 42.56 \\
    (3)~~MLP or CrossAttention & 39.25  & 37.68  & 42.89  \\
    (4)~~MLP + Concat & 40.24 & 38.23 & 44.39 \\ 
    (5)~~CrossAttention + Concat & 42.31 & 39.43 & 46.21 \\
    (6)~~CLIP as Q and LLM as KV & 43.69 & 42.82 & 49.56 \\
    (7)~~Ours & \textbf{47.34} & \textbf{45.35} & \textbf{55.39} \\
    \bottomrule
    \end{tabular}
    }
    \caption{Impact (\%) of the designed Cross-Adapter Module.}
    \label{tab:ab-design}
\end{table}

\subsubsection{Impact of Different LLMs}
The analysis encompasses a comparative evaluation between base SD1.5 and the enhancements achieved through the integration of Llama-2/7B, Llama-2/13B, and T5-XL. As depicted in \cref{tab:ab-llm}, the inclusion of any LLM improves upon the performance of SD1.5. Importantly, Llama-v2/13B outperforms Llama-v2/7B, demonstrating that LLMs with greater capacity excel in extracting more nuanced semantic embeddings. Furthermore, when compared to decoder-only LLMs, T5-XL encoder demonstrates advantages in semantic comprehension, confirming its superior suitability for enhancing text-to-image generation.

\subsection{Further Analysis}

\begin{table}[!t]
    \centering
    \resizebox{\linewidth}{!}{
    \begin{tabular}{lcc|c}
    \toprule
    \small Method & \small \#Images (M $\downarrow$) & \small \#GPU days ($\downarrow$) & \small Performance ($\uparrow$) \\
    \midrule
    PixArt-$\alpha$ \cite{chen2023pixart} & 25 & 753 & 68.86 \\
    ParaDiffusion \cite{wu2023paragraph} & 500 & 392 & - \\ 
    ELLA$_{SDXL}$ \cite{ella} & 30 & 112 & 72.60 \\
    \midrule
    LLM4GEN$_{SDXL}$ & 10 & 50 & \textbf{76.59} \\
    \bottomrule
    \end{tabular}
    }    
    \caption{Training resources comparison, including the scale of training data and computing cost.}
    \label{tab:training-resources}
\end{table}

\begin{figure}[!t]
    \centering
    \includegraphics[width=0.85\linewidth]{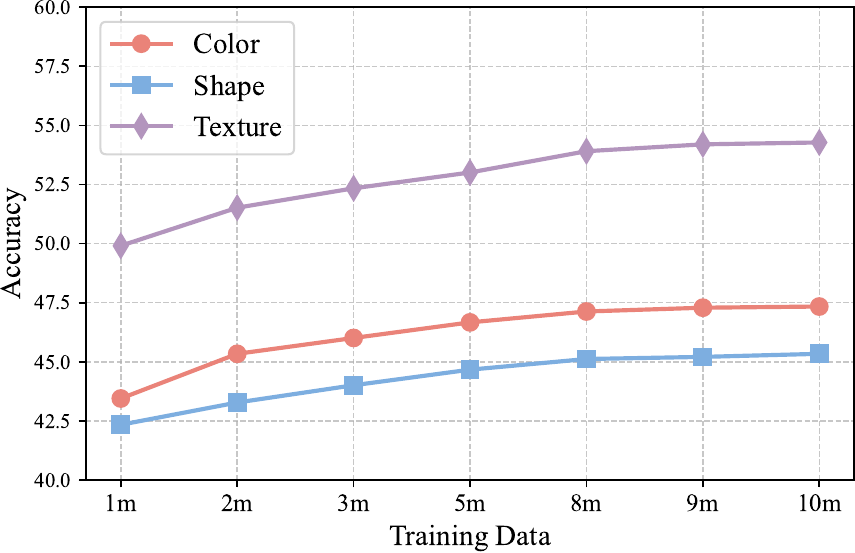}
    \caption{Training data scaling analysis.}
    \vspace{-2mm}
    \label{fig: training data}
\end{figure}

\begin{figure}[!t]
    \centering
    \includegraphics[width=\linewidth]{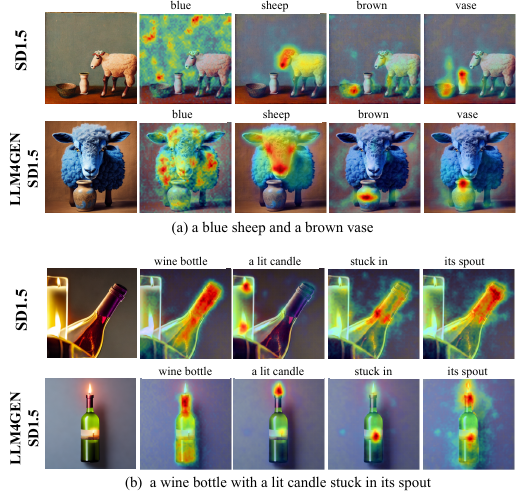}
    \caption{Cross-attention visualization \cite{tang2022daam} for two generated images. The two rows are SD1.5 and LLM4GEN$_{SD1.5}$, respectively.}
    \label{fig:visualization}
\end{figure}

\subsubsection{Scaling Analysis} As illustrated in \cref{fig: training data}, we conduct an extensive analysis of the scalability of our proposed LLM4GEN model with respect to training data. The results conclusively demonstrate that as the quantity of training data increases, the performance of our model exhibits consistent and significant growth, thereby confirming its scalability. However, increasing the dataset scale from 5M to 10M resulted in minimal performance improvement on the generated images. Consequently, we use 10M text-image pairs for training our LLM4GEN.

\subsubsection{Training Efficiency} When evaluating the effectiveness of integrating LLMs into text-to-image generation models, LLM4GEN$_{SDXL}$ stands out for its remarkable efficiency and performance. LLM4GEN achieves significant reductions in both training data requirements and computational costs. It utilizes only 10 million data, a 66\% reduction compared to ELLA, and demands merely 50 GPU days for training, drastically lower than PixArt-$\alpha$ (25 million data, 753 GPU days) and ParaDiffusion (500 million data, 392 GPU days). Despite this, LLM4GEN$_{SDXL}$ achieves a superior color metric performance of 73.29\%. This notable difference underscores LLM4GEN's ability to substantially reduce both training data and computational costs while establishing a new standard for performance efficiency.

\subsubsection{Cross-attention Visualization.} \cref{fig:visualization} shows the cross-attention visualization of SD1.5 and LLM4GEN$_{SD1.5}$, respectively. The heatmaps reveal that our proposed LLM4GEN method demonstrates a superior ability to capture relationships between attributes, such as "blue" and "sheep," as illustrated in \cref{fig:visualization}~(a). We attribute this enhanced capability to the increased semantic richness afforded by the robust representations of LLMs.

More visualization and experimental results are shown in the Appendix.

%% file: sections/conclusion.tex
\section{Conclusion}
In this paper, we propose LLM4GEN, an end-to-end text-to-image generation framework. Specifically, we design an efficient Cross-Adapter Module to leverage the powerful representation of LLMs, thereby enhancing the original text representation of diffusion models. Despite using fewer training data and computational resources, LLM4GEN outperforms current state-of-the-art text-to-image diffusion models in sample quality and image-text alignment. To optimize consistency in entity-attribute relationships of generated images, we design an entity-guided regularization loss. Additionally, we introduce the DensePrompts benchmark to promote the generation of images with dense information and provide a comprehensive evaluation framework. Extensive experiments have shown that our proposed method achieves competitive performance.

%% file: main.bbl
\begin{thebibliography}{42}
\providecommand{\natexlab}[1]{#1}

\bibitem[{Achiam et~al.(2023)Achiam, Adler, Agarwal, Ahmad, Akkaya, Aleman, Almeida, Altenschmidt, Altman, Anadkat et~al.}]{gpt-4}
Achiam, J.; Adler, S.; Agarwal, S.; Ahmad, L.; Akkaya, I.; Aleman, F.~L.; Almeida, D.; Altenschmidt, J.; Altman, S.; Anadkat, S.; et~al. 2023.
\newblock Gpt-4 technical report.
\newblock \emph{arXiv preprint arXiv:2303.08774}.

\bibitem[{Bai et~al.(2023)Bai, Bai, Yang, Wang, Tan, Wang, Lin, Zhou, and Zhou}]{bai2023qwen}
Bai, J.; Bai, S.; Yang, S.; Wang, S.; Tan, S.; Wang, P.; Lin, J.; Zhou, C.; and Zhou, J. 2023.
\newblock Qwen-vl: A frontier large vision-language model with versatile abilities.
\newblock \emph{arXiv preprint arXiv:2308.12966}.

\bibitem[{Betker et~al.(2023)Betker, Goh, Jing, Brooks, Wang, Li, Ouyang, Zhuang, Lee, Guo et~al.}]{betker2023improving}
Betker, J.; Goh, G.; Jing, L.; Brooks, T.; Wang, J.; Li, L.; Ouyang, L.; Zhuang, J.; Lee, J.; Guo, Y.; et~al. 2023.
\newblock Improving image generation with better captions.
\newblock \emph{Computer Science. https://cdn.openai.com/papers/dall-e-3.pdf}.

\bibitem[{Brown et~al.(2020)Brown, Mann, Ryder, Subbiah, Kaplan, Dhariwal, Neelakantan, Shyam, Sastry, Askell, Agarwal, Herbert-Voss, Krueger, Henighan, Child, Ramesh, Ziegler, Wu, Winter, Hesse, Chen, Sigler, Litwin, Gray, Chess, Clark, Berner, McCandlish, Radford, Sutskever, and Amodei}]{brown2020language}
Brown, T.~B.; Mann, B.; Ryder, N.; Subbiah, M.; Kaplan, J.; Dhariwal, P.; Neelakantan, A.; Shyam, P.; Sastry, G.; Askell, A.; Agarwal, S.; Herbert-Voss, A.; Krueger, G.; Henighan, T.; Child, R.; Ramesh, A.; Ziegler, D.~M.; Wu, J.; Winter, C.; Hesse, C.; Chen, M.; Sigler, E.; Litwin, M.; Gray, S.; Chess, B.; Clark, J.; Berner, C.; McCandlish, S.; Radford, A.; Sutskever, I.; and Amodei, D. 2020.
\newblock Language models are few-shot learners.

\bibitem[{Chang et~al.(2023)Chang, Wang, Wang, Wu, Yang, Zhu, Chen, Yi, Wang, Wang et~al.}]{chang2023survey}
Chang, Y.; Wang, X.; Wang, J.; Wu, Y.; Yang, L.; Zhu, K.; Chen, H.; Yi, X.; Wang, C.; Wang, Y.; et~al. 2023.
\newblock A survey on evaluation of large language models.
\newblock \emph{ACM Transactions on Intelligent Systems and Technology}.

\bibitem[{Chefer et~al.(2023)Chefer, Alaluf, Vinker, Wolf, and Cohen-Or}]{chefer2023attend}
Chefer, H.; Alaluf, Y.; Vinker, Y.; Wolf, L.; and Cohen-Or, D. 2023.
\newblock Attend-and-excite: Attention-based semantic guidance for text-to-image diffusion models.
\newblock \emph{ACM TOG}, 42(4): 1--10.

\bibitem[{Chen et~al.(2023{\natexlab{a}})Chen, Pan, Yao, and Mei}]{chen2023controlstyle}
Chen, J.; Pan, Y.; Yao, T.; and Mei, T. 2023{\natexlab{a}}.
\newblock Controlstyle: Text-driven stylized image generation using diffusion priors.
\newblock In \emph{ACM MM}, 7540--7548.

\bibitem[{Chen et~al.(2023{\natexlab{b}})Chen, Yu, Ge, Yao, Xie, Wu, Wang, Kwok, Luo, Lu et~al.}]{chen2023pixart}
Chen, J.; Yu, J.; Ge, C.; Yao, L.; Xie, E.; Wu, Y.; Wang, Z.; Kwok, J.; Luo, P.; Lu, H.; et~al. 2023{\natexlab{b}}.
\newblock PixArt-$\alpha$: Fast Training of Diffusion Transformer for Photorealistic Text-to-Image Synthesis.
\newblock \emph{arXiv preprint arXiv:2310.00426}.

\bibitem[{Chowdhery et~al.(2022)Chowdhery, Narang, Devlin, Bosma, Mishra, Roberts, Barham, Chung, Sutton, Gehrmann et~al.}]{chowdhery2022palm}
Chowdhery, A.; Narang, S.; Devlin, J.; Bosma, M.; Mishra, G.; Roberts, A.; Barham, P.; Chung, H.~W.; Sutton, C.; Gehrmann, S.; et~al. 2022.
\newblock Palm: Scaling language modeling with pathways.
\newblock \emph{arXiv preprint arXiv:2204.02311}.

\bibitem[{Feng et~al.(2022)Feng, He, Fu, Jampani, Akula, Narayana, Basu, Wang, and Wang}]{feng2022training}
Feng, W.; He, X.; Fu, T.-J.; Jampani, V.; Akula, A.~R.; Narayana, P.; Basu, S.; Wang, X.~E.; and Wang, W.~Y. 2022.
\newblock Training-Free Structured Diffusion Guidance for Compositional Text-to-Image Synthesis.
\newblock In \emph{ICLR}.

\bibitem[{Hu et~al.(2024)Hu, Wang, Fang, Fu, Cheng, and Yu}]{ella}
Hu, X.; Wang, R.; Fang, Y.; Fu, B.; Cheng, P.; and Yu, G. 2024.
\newblock ELLA: Equip Diffusion Models with LLM for Enhanced Semantic Alignment.
\newblock \emph{arXiv preprint arXiv:2403.05135}.

\bibitem[{Huang et~al.(2023)Huang, Sun, Xie, Li, and Liu}]{huang2023t2i}
Huang, K.; Sun, K.; Xie, E.; Li, Z.; and Liu, X. 2023.
\newblock T2I-CompBench: A Comprehensive Benchmark for Open-world Compositional Text-to-image Generation.
\newblock In \emph{ICCV}.

\bibitem[{Lian et~al.(2024)Lian, Li, Yala, and Darrell}]{lmd}
Lian, L.; Li, B.; Yala, A.; and Darrell, T. 2024.
\newblock {LLM}-grounded Diffusion: Enhancing Prompt Understanding of Text-to-Image Diffusion Models with Large Language Models.
\newblock \emph{Transactions on Machine Learning Research}.

\bibitem[{Lin et~al.(2014)Lin, Maire, Belongie, Hays, Perona, Ramanan, Doll{\'a}r, and Zitnick}]{microsoftcoco}
Lin, T.-Y.; Maire, M.; Belongie, S.; Hays, J.; Perona, P.; Ramanan, D.; Doll{\'a}r, P.; and Zitnick, C.~L. 2014.
\newblock Microsoft coco: Common objects in context.
\newblock In \emph{ECCV}.

\bibitem[{Liu et~al.(2022)Liu, Li, Du, Torralba, and Tenenbaum}]{liu2022compositional}
Liu, N.; Li, S.; Du, Y.; Torralba, A.; and Tenenbaum, J.~B. 2022.
\newblock Compositional visual generation with composable diffusion models.
\newblock In \emph{ECCV}, 423--439.

\bibitem[{Nichol et~al.(2022)Nichol, Dhariwal, Ramesh, Shyam, Mishkin, McGrew, Sutskever, and Chen}]{nichol2021glide}
Nichol, A.; Dhariwal, P.; Ramesh, A.; Shyam, P.; Mishkin, P.; McGrew, B.; Sutskever, I.; and Chen, M. 2022.
\newblock Glide: Towards photorealistic image generation and editing with text-guided diffusion models.
\newblock In \emph{ICML}.

\bibitem[{OpenAI(2023)}]{2023GPT4VisionSC}
OpenAI. 2023.
\newblock GPT-4V(ision) System Card.

\bibitem[{Pang et~al.(2024)Pang, Xie, Man, and Wang}]{pang2023frozen}
Pang, Z.; Xie, Z.; Man, Y.; and Wang, Y.-X. 2024.
\newblock Frozen transformers in language models are effective visual encoder layers.
\newblock In \emph{ICLR}.

\bibitem[{Podell et~al.(2023)Podell, English, Lacey, Blattmann, Dockhorn, M{\"u}ller, Penna, and Rombach}]{podell2023sdxl}
Podell, D.; English, Z.; Lacey, K.; Blattmann, A.; Dockhorn, T.; M{\"u}ller, J.; Penna, J.; and Rombach, R. 2023.
\newblock Sdxl: Improving latent diffusion models for high-resolution image synthesis.
\newblock \emph{arXiv preprint arXiv:2307.01952}.

\bibitem[{Radford et~al.(2021)Radford, Kim, Hallacy, Ramesh, Goh, Agarwal, Sastry, Askell, Mishkin, Clark et~al.}]{clip}
Radford, A.; Kim, J.~W.; Hallacy, C.; Ramesh, A.; Goh, G.; Agarwal, S.; Sastry, G.; Askell, A.; Mishkin, P.; Clark, J.; et~al. 2021.
\newblock Learning Transferable Visual Models from Natural Language Supervision.
\newblock In \emph{ICML}, 8748--8763.

\bibitem[{Raffel et~al.(2020)Raffel, Shazeer, Roberts, Lee, Narang, Matena, Zhou, Li, and Liu}]{raffel2020exploring}
Raffel, C.; Shazeer, N.; Roberts, A.; Lee, K.; Narang, S.; Matena, M.; Zhou, Y.; Li, W.; and Liu, P.~J. 2020.
\newblock Exploring the limits of transfer learning with a unified text-to-text transformer.
\newblock \emph{JMLR}, 21(1): 5485--5551.

\bibitem[{Ramesh et~al.(2022)Ramesh, Dhariwal, Nichol, Chu, and Chen}]{ramesh2022hierarchical}
Ramesh, A.; Dhariwal, P.; Nichol, A.; Chu, C.; and Chen, M. 2022.
\newblock Hierarchical text-conditional image generation with clip latents.
\newblock \emph{arXiv preprint arXiv:2204.06125}.

\bibitem[{Rombach et~al.(2022)Rombach, Blattmann, Lorenz, Esser, and Ommer}]{rombach2022high}
Rombach, R.; Blattmann, A.; Lorenz, D.; Esser, P.; and Ommer, B. 2022.
\newblock High-resolution image synthesis with latent diffusion models.
\newblock In \emph{CVPR}, 10684--10695.

\bibitem[{Ruiz et~al.(2023)Ruiz, Li, Jampani, Pritch, Rubinstein, and Aberman}]{ruiz2023dreambooth}
Ruiz, N.; Li, Y.; Jampani, V.; Pritch, Y.; Rubinstein, M.; and Aberman, K. 2023.
\newblock Dreambooth: Fine tuning text-to-image diffusion models for subject-driven generation.
\newblock In \emph{CVPR}, 22500--22510.

\bibitem[{Saharia et~al.(2022)Saharia, Chan, Saxena, Li, Whang, Denton, Ghasemipour, Gontijo~Lopes, Karagol~Ayan, Salimans et~al.}]{saharia2022photorealistic}
Saharia, C.; Chan, W.; Saxena, S.; Li, L.; Whang, J.; Denton, E.~L.; Ghasemipour, K.; Gontijo~Lopes, R.; Karagol~Ayan, B.; Salimans, T.; et~al. 2022.
\newblock Photorealistic text-to-image diffusion models with deep language understanding.
\newblock In \emph{NeurIPS}.

\bibitem[{Schuhmann()}]{improvedaestheticpredictor}
Schuhmann, C. ????
\newblock CLIP+MLP Aesthetic Score Predictor.
\newblock https://github.com/christophschuhmann/improved-aesthetic-predictor.

\bibitem[{Schuhmann et~al.(2021)Schuhmann, Vencu, Beaumont, Kaczmarczyk, Mullis, Katta, Coombes, Jitsev, and Komatsuzaki}]{laion}
Schuhmann, C.; Vencu, R.; Beaumont, R.; Kaczmarczyk, R.; Mullis, C.; Katta, A.; Coombes, T.; Jitsev, J.; and Komatsuzaki, A. 2021.
\newblock {LAION-400M:} Open Dataset of CLIP-Filtered 400 Million Image-Text Pairs.
\newblock \emph{CoRR}, abs/2111.02114.

\bibitem[{Song, Meng, and Ermon(2020)}]{ddim}
Song, J.; Meng, C.; and Ermon, S. 2020.
\newblock Denoising Diffusion Implicit Models.
\newblock \emph{CoRR}, abs/2010.02502.

\bibitem[{Song and Ermon(2019)}]{song2019generative}
Song, Y.; and Ermon, S. 2019.
\newblock Generative modeling by estimating gradients of the data distribution.

\bibitem[{Song and Ermon(2020)}]{song2020improved}
Song, Y.; and Ermon, S. 2020.
\newblock Improved techniques for training score-based generative models.

\bibitem[{Song et~al.(2020)Song, Sohl-Dickstein, Kingma, Kumar, Ermon, and Poole}]{song2020score}
Song, Y.; Sohl-Dickstein, J.; Kingma, D.~P.; Kumar, A.; Ermon, S.; and Poole, B. 2020.
\newblock Score-based generative modeling through stochastic differential equations.
\newblock \emph{arXiv preprint arXiv:2011.13456}.

\bibitem[{Sun et~al.(2024)Sun, Yu, Cui, Zhang, Zhang, Wang, Gao, Liu, Huang, and Wang}]{emu}
Sun, Q.; Yu, Q.; Cui, Y.; Zhang, F.; Zhang, X.; Wang, Y.; Gao, H.; Liu, J.; Huang, T.; and Wang, X. 2024.
\newblock Emu: Generative pretraining in multimodality.
\newblock In \emph{ICLR}.

\bibitem[{Tang et~al.(2022)Tang, Liu, Pandey, Jiang, Yang, Kumar, Stenetorp, Lin, and Ture}]{tang2022daam}
Tang, R.; Liu, L.; Pandey, A.; Jiang, Z.; Yang, G.; Kumar, K.; Stenetorp, P.; Lin, J.; and Ture, F. 2022.
\newblock What the daam: Interpreting stable diffusion using cross attention.
\newblock \emph{arXiv preprint arXiv:2210.04885}.

\bibitem[{Touvron et~al.(2023)Touvron, Martin, Stone, Albert, Almahairi, Babaei, Bashlykov, Batra, Bhargava, Bhosale et~al.}]{touvron2023llama}
Touvron, H.; Martin, L.; Stone, K.; Albert, P.; Almahairi, A.; Babaei, Y.; Bashlykov, N.; Batra, S.; Bhargava, P.; Bhosale, S.; et~al. 2023.
\newblock Llama 2: Open foundation and fine-tuned chat models.
\newblock \emph{arXiv preprint arXiv:2307.09288}.

\bibitem[{Wu et~al.(2023)Wu, Li, He, Shou, Shen, Cheng, Li, Gao, Zhang, and Wang}]{wu2023paragraph}
Wu, W.; Li, Z.; He, Y.; Shou, M.~Z.; Shen, C.; Cheng, L.; Li, Y.; Gao, T.; Zhang, D.; and Wang, Z. 2023.
\newblock Paragraph-to-image generation with information-enriched diffusion model.
\newblock \emph{arXiv preprint arXiv:2311.14284}.

\bibitem[{Yang et~al.(2024)Yang, Yu, Meng, Xu, Ermon, and Cui}]{RPG}
Yang, L.; Yu, Z.; Meng, C.; Xu, M.; Ermon, S.; and Cui, B. 2024.
\newblock Mastering text-to-image diffusion: Recaptioning, planning, and generating with multimodal llms.
\newblock \emph{arXiv preprint arXiv:2401.11708}.

\bibitem[{Yu et~al.(2022)Yu, Xu, Koh, Luong, Baid, Wang, Vasudevan, Ku, Yang, Ayan et~al.}]{yu2022scaling}
Yu, J.; Xu, Y.; Koh, J.~Y.; Luong, T.; Baid, G.; Wang, Z.; Vasudevan, V.; Ku, A.; Yang, Y.; Ayan, B.~K.; et~al. 2022.
\newblock Scaling autoregressive models for content-rich text-to-image generation.
\newblock \emph{arXiv preprint arXiv:2206.10789}.

\bibitem[{Zhang et~al.(2024)Zhang, Han, Zhou, Hu, Yan, Lu, Li, Gao, and Qiao}]{zhang2023llama}
Zhang, R.; Han, J.; Zhou, A.; Hu, X.; Yan, S.; Lu, P.; Li, H.; Gao, P.; and Qiao, Y. 2024.
\newblock Llama-adapter: Efficient fine-tuning of language models with zero-init attention.
\newblock In \emph{ICLR}.

\bibitem[{Zhang et~al.(2022)Zhang, Roller, Goyal, Artetxe, Chen, Chen, Dewan, Diab, Li, Lin, Mihaylov et~al.}]{zhang2022opt}
Zhang, S.; Roller, S.; Goyal, N.; Artetxe, M.; Chen, M.; Chen, S.; Dewan, C.; Diab, M.; Li, X.; Lin, X.~V.; Mihaylov, T.; et~al. 2022.
\newblock Opt: Open pre-trained transformer language models.
\newblock \emph{arXiv preprint arXiv:2205.01068}.

\bibitem[{Zhao et~al.(2024)Zhao, Hao, Zi, Xu, and Wong}]{bridge}
Zhao, S.; Hao, S.; Zi, B.; Xu, H.; and Wong, K.-Y.~K. 2024.
\newblock Bridging Different Language Models and Generative Vision Models for Text-to-Image Generation.
\newblock In \emph{ECCV}.

\bibitem[{Zhong et~al.(2023)Zhong, Huang, Wen, Qin, and Lin}]{suradapter}
Zhong, S.; Huang, Z.; Wen, W.; Qin, J.; and Lin, L. 2023.
\newblock SUR-adapter: Enhancing Text-to-Image Pre-trained Diffusion Models with Large Language Models.
\newblock arXiv:2305.05189.

\bibitem[{Zhu et~al.(2023)Zhu, Chen, Shen, Li, and Elhoseiny}]{zhu2023minigpt}
Zhu, D.; Chen, J.; Shen, X.; Li, X.; and Elhoseiny, M. 2023.
\newblock Minigpt-4: Enhancing vision-language understanding with advanced large language models.
\newblock \emph{arXiv preprint arXiv:2304.10592}.

\end{thebibliography}
